\pdfoutput=1

\documentclass[11pt]{article}

\usepackage[final]{acl}

\usepackage{times}
\usepackage{latexsym}

\usepackage[T1]{fontenc}

\usepackage[utf8]{inputenc}

\usepackage{microtype}

\usepackage{inconsolata}

\usepackage{graphicx}
\usepackage[export]{adjustbox}

\usepackage{booktabs}

\usepackage{xcolor}

\usepackage{algorithm}
\usepackage{algorithmic}
\usepackage{amsmath}
\usepackage{newfloat}
\usepackage{listings}

\title{How do Transformer Embeddings Represent Compositions? \newline A Functional Analysis}

\author{
Aishik Nagar\textsuperscript{1} \quad 
Ishaan Rawal\textsuperscript{2,3,4} \quad 
Mansi Dhanania\textsuperscript{2,3,5} \quad 
Cheston Tan\textsuperscript{2,3} \quad
\\
\\
\textsuperscript{1}ASUS Intelligent Cloud Services \\
\textsuperscript{2}Institute of High Performance Computing (IHPC), Agency for Science, Technology and Research (A*STAR) \\
\textsuperscript{3}Center for Frontier AI Research (CFAR), Agency for Science, Technology and Research (A*STAR) \\
\textsuperscript{4}Texas A\&M University \\
\textsuperscript{5}McGill University \\
Correspondence to: \texttt{aishiknagar@gmail.com}
}

\begin{document}
\maketitle
\begin{abstract}
Compositionality is a key aspect of human intelligence, essential for reasoning and generalization. While transformer-based models have become the de facto standard for many language modeling tasks, little is known about how they represent compound words, and whether these representations are compositional. In this study, we test compositionality in Mistral, OpenAI Large, and Google embedding models, and compare them with BERT. First, we evaluate compositionality in the representations by examining six diverse models of compositionality (addition, multiplication, dilation, regression, etc.). We find that ridge regression, albeit linear, best accounts for compositionality. Surprisingly, we find that the classic vector addition model performs almost as well as any other model. Next, we verify that most embedding models are highly compositional, while BERT shows much poorer compositionality. We verify and visualize our findings with a synthetic dataset consisting of fully transparent adjective-noun compositions. Overall, we present a thorough investigation of compositionality.
\end{abstract}

\section{Introduction}

\begin{figure}
\vskip 0.2in
\begin{center}
\centerline{\includegraphics[scale=0.5]{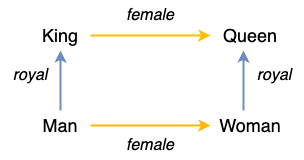}}
\caption{Linear concept transformation functions (\textit{female} and \textit{royal}) used to transform the representations of nouns \cite{ethayarajh-etal-2019-towards}.}
\label{fig:VectorImage}
\end{center}
\vskip -0.2in
\end{figure}

Large Language Models (LLMs) have emerged as the de facto standard for language generation and modeling tasks \cite{LLMSurvey}, largely due to their superior generalization abilities compared to earlier models like BERT \cite{devlin-etal-2019-bert}. One key factor contributing to this generalizability is \textbf{compositionality -- the ability to form new concepts by combining known ones}~\cite{fodor2002compositionality}. However, testing the compositionality of these models is challenging. Previous works on transformer models have mostly relied on performance on benchmarks as proxies for compositional generalization \cite{johnson2017clevr, yu2022crepe}, but such numbers can be often misleading \cite{pmlr-v235-rawal24a}. Thus, representation analysis, like that in Fig. \ref{fig:VectorImage}, can provide insights about the inner workings of a model. Thus, we conduct representation-centric analyses to verify the compositionality of transformer models.

While a substantial body of literature exists on the functional compositionality of Vector Space Models (VSMs)~\cite{mitchell2008vector, baroni-zamparelli-2010-nouns}, such understanding of transformer models, which also rely on the distributional hypothesis \cite{firth1957synopsis}, remains limited. Additionally, exploring differences in how current models comprehend compositionality is crucial. To address these, we adopt a functional approach to investigate the compositionality of transformer-based embedding models. We focus on embedding models, due to their widespread use in applications like Retrieval-Augmented Generation~\cite{lewis2020retrieval} and vector databases~\cite{han2023comprehensive}.
The primary goal of this work is specifically to analyze how transformer models represent constitutuents of compound words, and how these token-level representations relate to the embedding generated for the entire compound word, taken as context. Transformer models excel at generating contextual embeddings, which differ fundamentally from static embeddings traditionally used in vector space models (VSMs). This context-dependent embedding generation allows transformer models to represent higher-order semantic features, such as semantic dominance and transparency, which emerge naturally in language usage. For example, the model might have learned that the term "Red Cross," although literally composed of "red" and "cross," frequently carries an organization-like meaning distinct from a purely additive combination of its constituents, resulting in a lower similarity between their consituent representations and the compound, regardless of the method of tokenization.
While it may seem intuitive that transformer models represent compound words as weighted averages of their constituent parts, the non-linear and multi-layered nature of these models makes composition far less predictable \cite{templeton2024scaling}. Tokenized compounds (e.g., “blue” + “tooth”) are not always interpreted in a compositional way, and may instead be mapped to distinct, non-compositional meanings based on learned context \cite{haber-poesio-2024-polysemy, grindrod2024transformers}. This ambiguity—especially for polysemous or rare compounds—motivates a closer examination of whether transformer embeddings respect compositionality in practice.

Compound words are a significant aspect of natural language. In English, compound words account for approximately 4\% of the total vocabulary in large corpora like the British National Corpus \cite{biber1999longman}.
Language models often fail to generalize to novel compounds or properly encode existing ones \cite{cordeiro2016predicting} and if a model struggles with word-level composition, it may face challenges with more complex structures \cite{lake2017building}. 
Understanding how models handle lexical composition is therefore foundational for scaling up to phrases and sentences. 
Adjective-noun compositions and compound words have been widely studied as tests of compositionality~\cite{reddy2011empirical, baroni-zamparelli-2010-nouns}.
They allow studying semantic composition in a controlled setting, minimizing syntactic variability. 
Although seemingly simple, two-word compounds and compositions can present significant challenges. 
For example, \textit{ladybird} is a beetle unrelated to female birds. \textit{Red ball} refers to a red-colored ball, whereas \textit{chocolate ball} refers to a ball made of chocolate. 
Moreover, humans often create novel compositions, such as \textit{deepfake}, which first appeared in 2018. 
As such, we focus on two-word compounds and adjective-noun compositions.

We extend the study of functional compositionality in VSMs to embedding models, by testing six compositionality models for combining constituent word embeddings to generate the embedding of their composition. We find that \textit{(i)} ridge regression can effectively explain compositionality in these models, and \textit{(ii)} current models are significantly more compositional than BERT. We also test the generalizability of our method on novel compound words created through data augmentation. Finally, we test the properties of the models on the LADEC dataset and on our Simple Adjective-Noun Compositions (SANC) dataset to \textit{(iii}) investigate and visualize the underlying mechanics.

\section{Related Works}


\citet{coil2023chocolate} questioned whether language models possess the ability to understand noun compounds. They inferred that GPT-3 was able to correctly understand the implicit relation between two nouns in compound words such as \textit{chocolate crocodile}, and conclude that it denotes a \textit{crocodile-shaped chocolate} due to the large amount of data it has access to through the web corpus. 

Another study, by \citet{buijtelaar2023psycholinguistic}, utilized human judgement to evaluate two psycholinguistic measures of compound semantic analysis, namely, lexeme meaning dominance and semantic transparency. This work demonstrated that BERT resembled human intuition in terms of semantic measures, especially when provided sufficient context to the data inputted.
 \citet{turney2010frequency} and \citet{bullinaria2012extracting}'s work on Distributional Semantic Models (DSMs) is discussed in appendix \ref{sec:related-work-continued}.

\textbf{Compositional Distributional Models.} \citet{mitchell2008vector} explored various compositional distributional models that derive vectors for phrases by composing the vectors of their constituent words. Their work on additive and multiplicative models provided a foundational understanding of how simple algebraic operations could approximate the meaning of phrases. \citet{baroni-zamparelli-2010-nouns} extended this by introducing the lexical function model, treating adjectives as functions mapping noun vectors to phrase vectors.


\textbf{Word Embeddings and Neural Networks.} The advent of word embeddings, such as Word2Vec \cite{mikolov2013efficient} and BERT \cite{bert}, provided dense, context-sensitive representations of words. These models have been instrumental in capturing subtle semantic nuances and have been widely used for various NLP tasks. Recent works on variational autoencoders for semantic analysis, such as~\citet{kingma2013auto}, have expanded understanding of embedding spaces by identifying latent dimensions representing semantic attributes like ``smiling'' and ``frowning''.

\textbf{Compositionality in Neural Language Models.} \citet{carvalho2022montague} proposed a methodology for measuring compositional behavior in contemporary neural language models, focusing on adjectival modifier phenomena in adjective-noun compositions. They introduced novel tests inspired by Montague semantics to evaluate whether neural embeddings can reflect theoretical compositional properties, such as intersectivity and non-subsectivity. Their findings indicated that while neural language models capture some aspects of these properties, they often fail to do so consistently, particularly in handling non-intersective adjectives.


While existing studies have shed light on the compositional abilities of both transformer and non-transformer models, they overlook the detailed exploration of latent embeddings and specific embedding models. By examining specific embedding models provided by Google, OpenAI, and Mistral alongside traditional embeddings like BERT, we provide a comprehensive analysis that uncovers nuanced semantic behaviors and enhances our understanding of embedding-driven compositionality.


\section{Testing Compositionality Models}

We test 6 compositionality models (both simple mathematical models and ones with learned parameters). These are tested on 4 common embeddings.

\subsection{Methods}

\subsubsection{Definitions}

For a given 2-word compound $c$, the first constituent word is $c_1$, while $c_2$ is the second constituent word. We define the embeddings of $c_1$, $c_2$ and $c$ to be vectors $u$, $v$ and $w$ respectively.

\subsubsection{Compositionality Models}

A model of compositionality is a function $f(u,v)= \hat{w}$, which produces a predicted embedding $\hat{w}$ by combining $u$ and $v$.

\begin{itemize}
    \item \textbf{Simple Addition:}
    \textcolor{black}{~~$f(u,v)=u+v$ } \\
    Straightforward sum of the embeddings of the constituent words, as the simplest mathematical relation possible between two vectors~\cite{park2023linear}.

    \item \textbf{Weighted Addition:}
    \textcolor{black}{~~$f(u,v)=\alpha u + \beta v $ } \\
    Following~\citet{mitchell2008vector, mitchell2009language, mitchell2010composition}, embeddings are combined using a weighted sum.

    \item \textbf{Multiplication:}
    \textcolor{black}{~~$f(u,v) = u \circ v $} \\
    Element-wise multiplication, capturing feature intersections where both constituent embeddings have high values~\citep{baroni-zamparelli-2010-nouns}.

    \item \textbf{Dilation:}
    \textcolor{black}{~~$f(u,v) = (u \cdot u)v + (\lambda - 1)(u \cdot v)u$ } \\
    Used by \citet{mitchell2008vector, mitchell2009language, mitchell2010composition} and \citet{baroni-zamparelli-2010-nouns}, dilation involves stretching one vector towards another by dilation parameter $\lambda$.

    \item \textbf{Ridge Regression:}
    \\
    The target values $w$ are regressed against the concatenation of $u$ and $v$~\cite{baroni-zamparelli-2010-nouns, vecchi2017spicy}, with parameter $\alpha=1.0$. Linear regression gave very similar results. 

    \item \textbf{Multi-Layer Perceptron (MLP):}
    \\
    A simple MLP with two hidden layers (256 neurons and 128 neurons respectively), ReLU activation, and cosine embedding loss function. The train-test split was 80-20.

\end{itemize}

\subsubsection{Optimization}
For the Weighted Addition and Dilation models, parameters were tuned using grid search and gradient descent. Since results using grid search were consistently worse than gradient descent, we omit them. More details are in Supplementary Materials.

\subsection{Experimental Setup}


\subsubsection{LADEC Dataset}
The LADEC dataset \cite{gagne2019ladec} is a set of 8,956 compound words and their constituent words. For each compound word, semantic ratings and other psycholinguistic markers are provided. These markers include measures of semantic transparency, which reflect how much of the meaning of the compound can be predicted from its parts, and meaning retention, which indicates how much each constituent retains its meaning within the compound. These ratings were obtained from human participants, who were asked to evaluate compounds based on specific criteria in controlled laboratory settings. Examples of compounds in LADEC are: bookstore, brainstorm, copycat, dumbbell, hamstring, hotdog, toothache, turtleneck and waterfall.


\subsubsection{LADEC--NC Dataset}

We created an extension of LADEC to assess how embedding models represent novel out-of-dictionary compounds words, providing insights into robustness and generalization. For example, a compound word like `deepfake' only came into existence in the era of generative vision models, but people easily grasped the gist of this novel (back then) compound from its constituent words `deep' (from \textit{deep learning}) and `fake'. \textbf{To what degree do the compositional capabilities of embedding models extend to novel compounds?} 
In the absence of human ratings, it can be postulated that these novel compounds have moderate to high semantic transparency -- human interpretations of terms like ``zoodough'' would naturally combine meanings of their constituents (e.g."dough used in a zoo") in the absence of contextual cues or prior lexical knowledge indicating that the compound carries an idiomatic or non-compositional meaning.

Starting with the compounds in LADEC, we generated lists of unique $c_1$ (first constituent word) and $c_2$ (second constituent word), e.g. `deep' and `fake' respectively. We then generated all pairwise combinations (i.e. compounds) from $c_1$ and $c_2$, filtering out any existing words to ensure novelty. Finally, 10,000 of these were randomly selected to form the \textbf{LADEC -- Novel Compounds (LADEC--NC)} dataset. Example novel compounds include: widowrise, throwlooms, hemnut and zoodough.

\subsubsection{Embedding Models}

We examine four different embedding models [dimensionality in brackets]: Google\footnote{\url{https://cloud.google.com/vertex-ai/generative-ai/docs/embeddings/get-text-embeddings}} [768], Mistral\footnote{\url{https://docs.mistral.ai/capabilities/embeddings/}} [1024], OpenAI Large\footnote{\url{https://platform.openai.com/docs/guides/embeddings/frequently-asked-questions}} [3072], and BERT [768] \citep{bert}. For BERT, we consider output embeddings corresponding to the \texttt{CLS} token 12th layer, because it holds maximum semantic information~\cite{mohebbi2021exploring} while there are proprietary APIs to obtain the embeddings for the other models.
Our selection of the primary models was motivated by their widespread adoption in industry and academic research (for example, the usage of BERT in metrics like BERTScore \cite{zhang2019bertscore} and the usage of the proprietary models in applications such as generating document representations and tasks such as retrieval-augmented-generation in LLMs). Resource and space constraints necessitated selecting a representative subset of models that clearly illustrate architectural differences (bidirectional vs. encoder-decoder embeddings) while being widely used for a diverse range of tasks which rely on the models to generate compositional and meaningful representations from text.

\subsubsection{Metrics}

To evaluate the different compositionality methods, we compute the cosine similarity between the predicted ($\hat{w}$) and actual ($w$) embeddings for each compound. These cosine similarities are then compared against a baseline distribution of cosine similarities between 1,000 random pairs of LADEC compounds (e.g. between `bookstore ' and `seashell'; between `hamstring' and `turtleneck').
If random compound embeddings showed similarities comparable to actual compounds and their constituents, tasks relying on embedding uniqueness -- such as document retrieval or text similarity (e.g. BERTScore \cite{zhang2019bertscore}) -- would be infeasible. Thus, comparing actual compound similarities against a random baseline confirms whether embeddings encode meaningful semantic distinctions or are merely noise.
The purpose of the random baseline is not to capture variance across transparency levels, but rather to assess whether compound–constituent pairs lie meaningfully apart from random pairs at the distribution level as a sanity check for our approach.

To quantify the difference between a method's distribution and the baseline distribution, we calculate the Jensen-Shannon (JS) Divergence \citep{lin1991divergence}. This measure of dissimilarity between two distributions is symmetric, and always gives finite values between 0 and 1. Two distributions are similar when their JS Divergence is close to 0. In this work, a larger divergence indicates that a method better captures semantic compositionality, i.e. larger difference from the baseline distribution.

\subsubsection{Visualization}

We use Kernel Density Estimation (KDE) plots to visualize the distribution of cosine similarities. Peaks in the plots indicate typical similarity levels, while the curve spread shows variability.

\subsection{Results}
\label{LADEC_results}

From Figure~\ref{fig:main_results_plots} and Table~\ref{table:main_results_table}, the most striking results at first glance are that:~~\textbf{(i)} the Multiplication model does extremely poorly across all embeddings, with the blue distributions lower than the baselines in red;~~\textbf{(ii)} BERT does not performed well, with large overlaps between the blue and red distributions. Detailed analyses is below and \textbf{Appendix \ref{sec:deeper-analysis}}.

\subsubsection{How do modern transformer-based embedding models compare to BERT?}   From Figure~\ref{fig:main_results_plots}, BERT is much less compositional than Google, Mistral and OpenAI Large. Apart from the Multiplication model (which performs poorly across all embeddings), all models of compositionality show large separations between the blue and red plots for all embedding models other than BERT. This is also reflected quantitatively in Table~\ref{table:main_results_table}; for all compositionality models, BERT has much lower JS Divergence than all others.

\subsubsection{Which compositionality model best predicts compound embeddings?}   From Table~\ref{table:main_results_table}, for all embedding models, Ridge Regression results in the highest JS Divergence values (by a small margin) and highest cosine similarities. MLP produces JS Divergence values close to those of Ridge Regression, but produces generally lower cosine similarities. Given that Ridge Regression is a linear model, it is somewhat surprising that it outperforms the non-linear MLP.

\subsubsection{How does the classic simple vector addition model perform?}   For the modern transformer-based embeddings (Google, Mistral and OpenAI Large), this simplest model of compositionality performs surprisingly well. The JS Divergence values are only marginally lower than those for Ridge Regression and NLP, while the mean cosine similarities are near (89\%) to those for Ridge Regression: 0.772 vs. 0.868 (average over 3 embedding models). Interestingly, the performance of Simple Addition is virtually the same as Weighted Addition and Dilation. Overall, this \underline{classic simple model holds up surprisingly well} for predicting modern compound embeddings.

\subsubsection{Does compositionality arise from architecture or training data?}   
\begin{figure*}[ht!]
\centering
\setlength{\tabcolsep}{1pt} 
\renewcommand{\arraystretch}{0.1} 
\begin{tabular}{r@{}c@{}c@{}c@{}c@{}}
  & \rotatebox{0}{\parbox{2.2cm}{\centering Google}} & \rotatebox{0}{\parbox{2.2cm}{\centering Mistral}} & \rotatebox{0}{\parbox{2.2cm}{\centering OpenAI Large}} & \rotatebox{0}{\parbox{2.2cm}{\centering BERT}} \\
  Simple Addition & 
  \includegraphics[width=0.18\linewidth,valign=m]{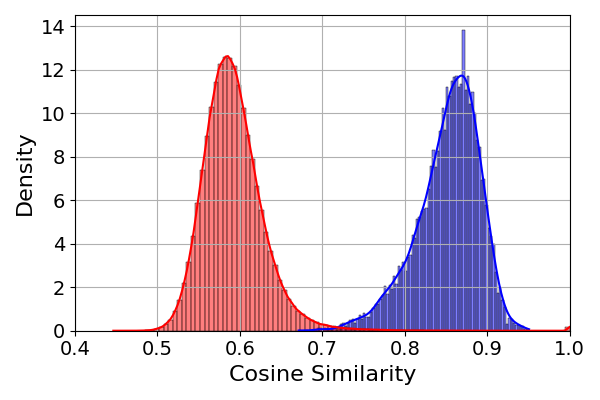} & 
  \includegraphics[width=0.18\linewidth,valign=m]{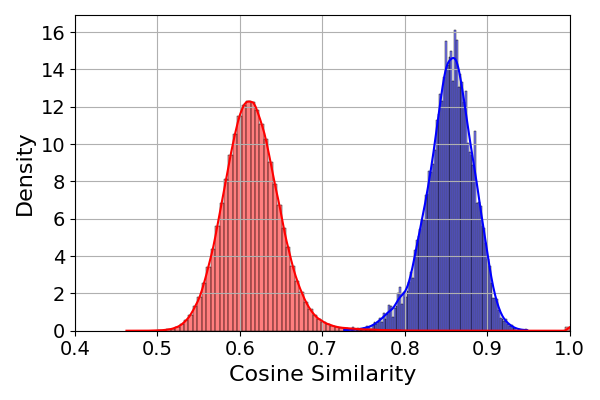} & 
  \includegraphics[width=0.18\linewidth,valign=m]{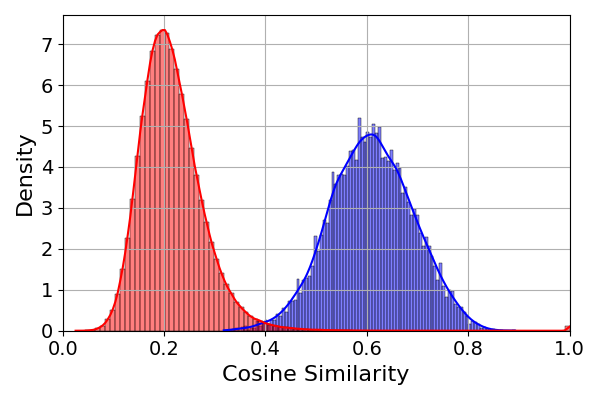} & 
  \includegraphics[width=0.18\linewidth,valign=m]{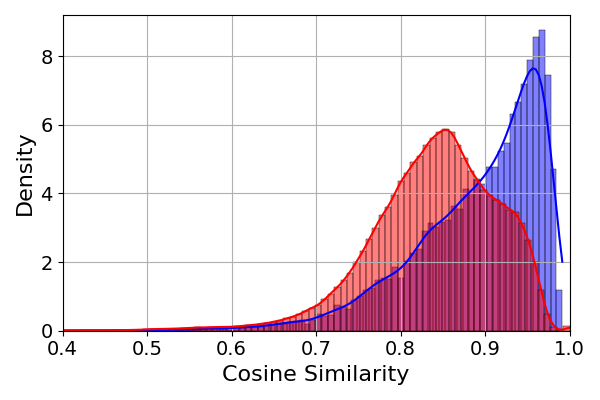} \\
  
  Weighted Addition & 
  \includegraphics[width=0.18\linewidth,valign=m]{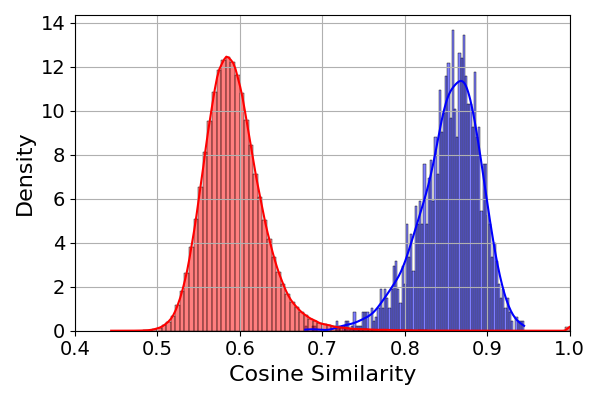} & 
  \includegraphics[width=0.18\linewidth,valign=m]{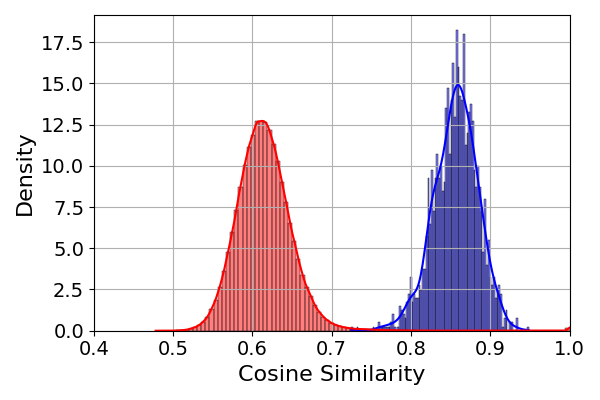} & 
  \includegraphics[width=0.18\linewidth,valign=m]{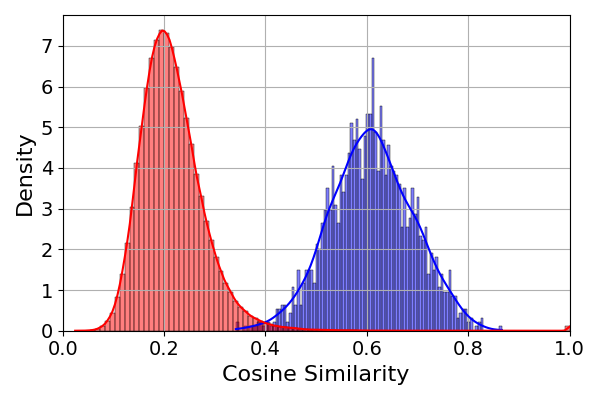} & 
  \includegraphics[width=0.18\linewidth,valign=m]{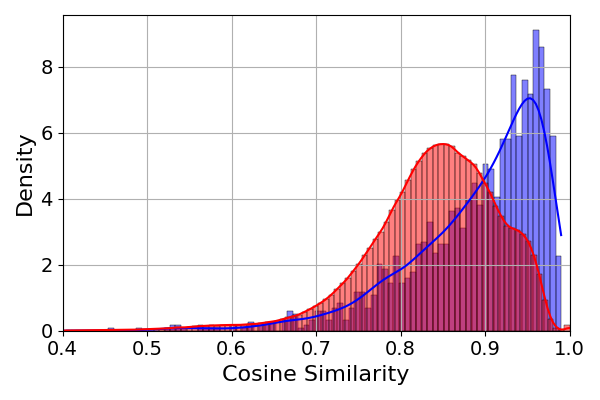} \\
  
  Multiplication & 
  \includegraphics[width=0.18\linewidth,valign=m]{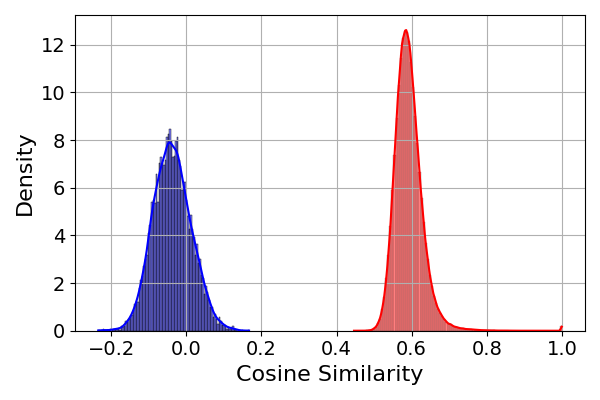} & 
  \includegraphics[width=0.18\linewidth,valign=m]{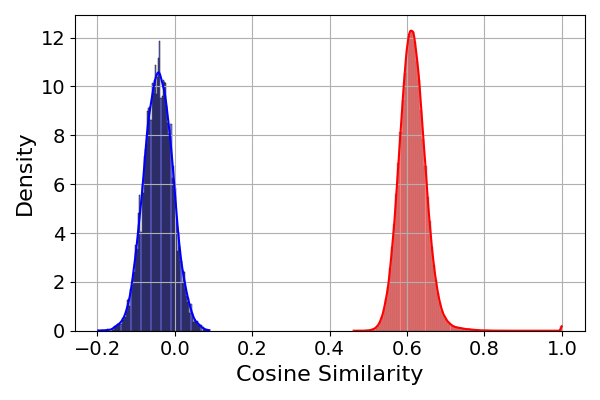} & 
  \includegraphics[width=0.18\linewidth,valign=m]{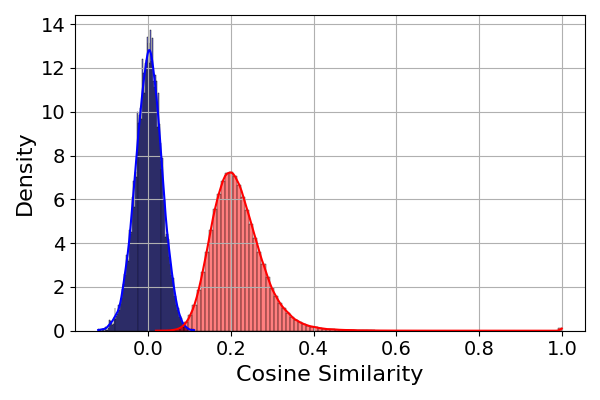} & 
  \includegraphics[width=0.18\linewidth,valign=m]{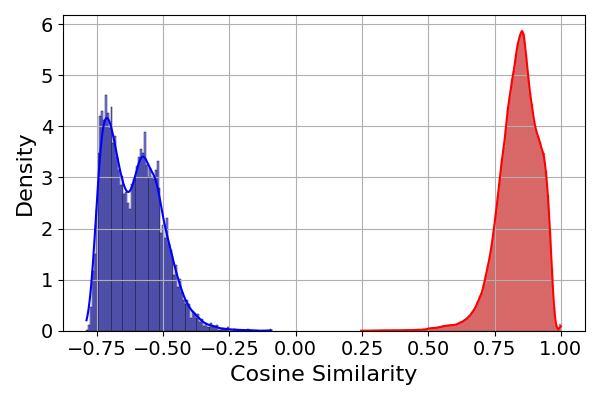} \\
  
  Dilation & 
  \includegraphics[width=0.18\linewidth,valign=m]{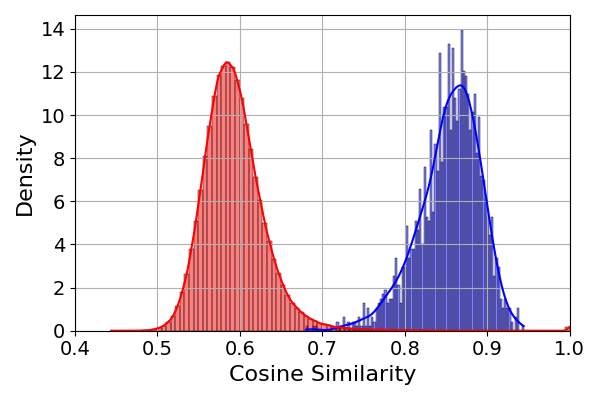} & 
  \includegraphics[width=0.18\linewidth,valign=m]{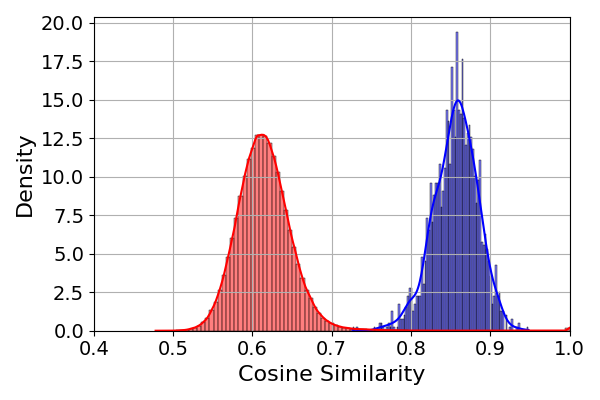} & 
  \includegraphics[width=0.18\linewidth,valign=m]{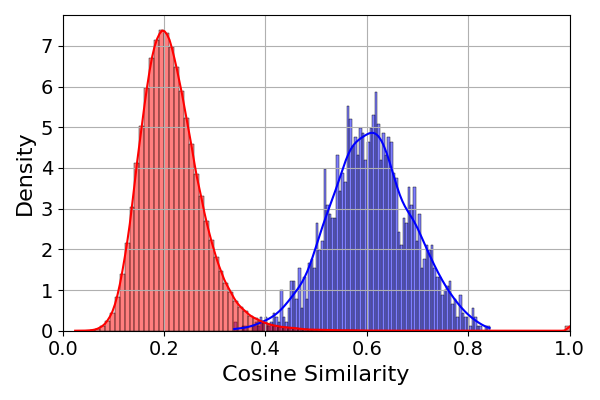} & 
  \includegraphics[width=0.18\linewidth,valign=m]{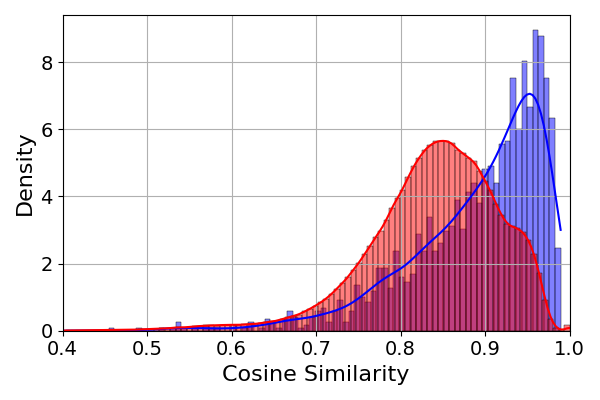} \\
  
  Ridge Reg. & 
  \includegraphics[width=0.18\linewidth,valign=m]{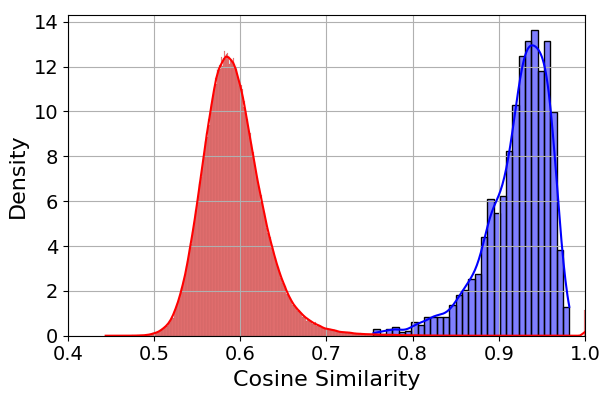} & 
  \includegraphics[width=0.18\linewidth,valign=m]{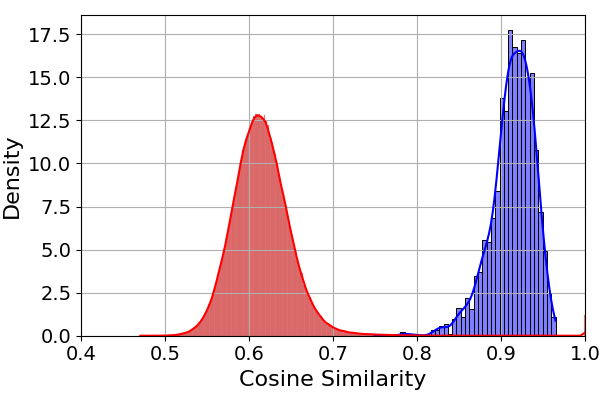} & 
  \includegraphics[width=0.18\linewidth,valign=m]{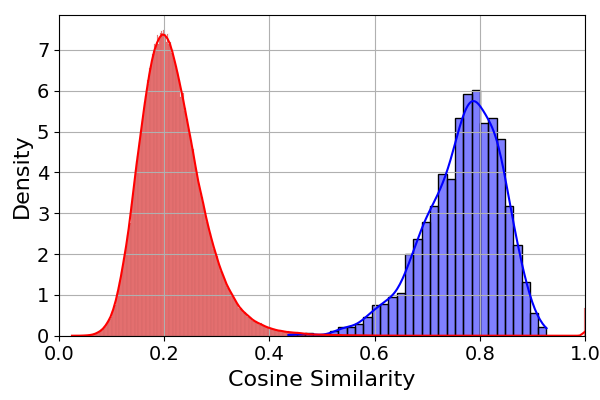} & 
  \includegraphics[width=0.18\linewidth,valign=m]{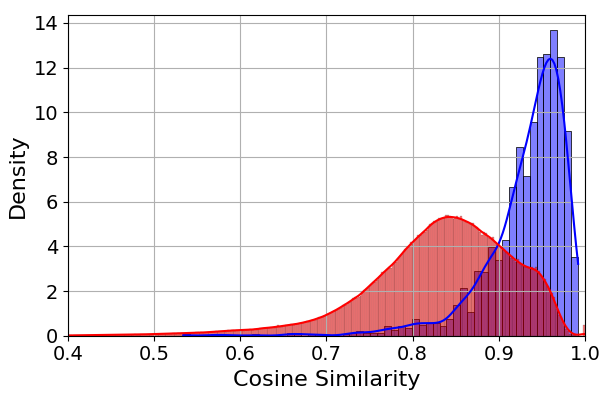} \\
  
  MLP & 
  \includegraphics[width=0.18\linewidth,valign=m]{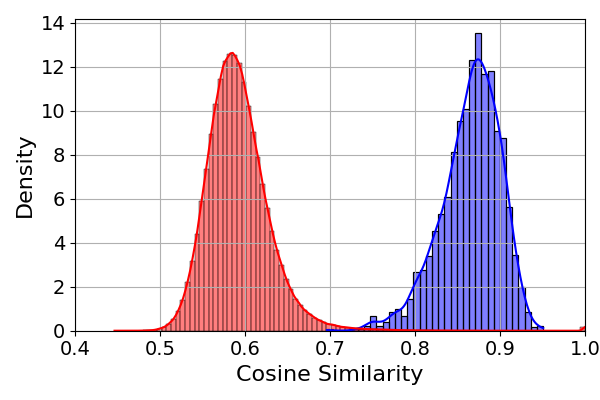} & 
  \includegraphics[width=0.18\linewidth,valign=m]{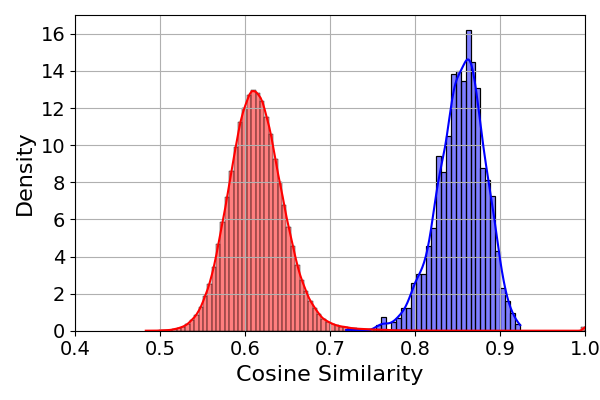} & 
  \includegraphics[width=0.18\linewidth,valign=m]{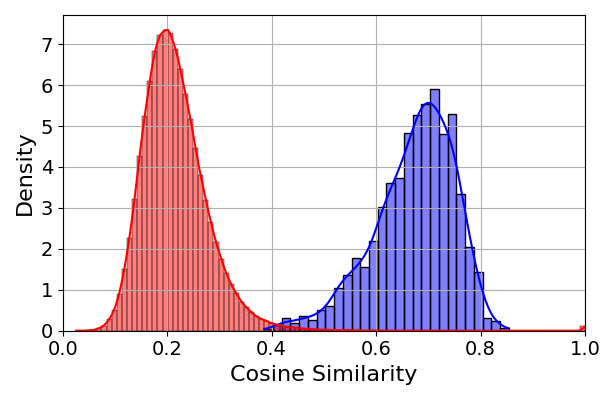} & 
  \includegraphics[width=0.18\linewidth,valign=m]{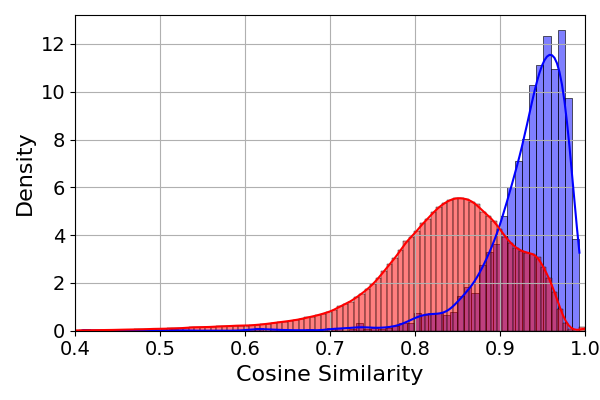} \\
  
\end{tabular}
\caption{Comparison across compositionality models (rows) and embedding models (columns). \textcolor{blue}{Blue plots} are the KDEs of cosine similarity values between the predicted ($\hat{w}$) and ground truth ($w$) embeddings, while the \textcolor{red}{red plots} are the baseline distributions (KDEs of cosine similarities between embeddings of random pairs of compounds).}
\label{fig:main_results_plots}
\end{figure*}

From Figure~\ref{fig:main_results_plots} and Table~\ref{table:main_results_table}, a number of embeddings and compositionality models show good results, with both high JS Divergence and cosine similarities. Could the existence of compounds words and their constituents (e.g. `deepfake', `deep' and `fake') in training corpora give rise to compositionality? \textbf{In other words, is compositionality innate or learnt?} From Table~\ref{table:metrics_NC}, the metrics for LADEC versus LADEC-NC are very similar. This means that embeddings of never-seen-before compounds (e.g. `zoodough') can be predicted from the embeddings of their constituents (`zoo' and 'dough') to the same extent as existing compounds. The results suggest that compositionality arises from the architecture of embedding models, rather than their training data or recipe.

\begin{table}[htbp]
\centering
\begin{tabular}{|l|l|r|r|}
\hline
\textbf{Embed.} & \textbf{Composition} & \textbf{Cosine} & \textbf{JS}   \\
\textbf{Model}     & \textbf{Model}            & \textbf{Sim.}   & \textbf{Div.} \\ \hline
\hline
Google        & Simple Addition   &                     0.852   &                     0.824  \\ \hline
Google        & Weighted Addition &                     0.854   &                     0.824  \\ \hline
Google        & Multiplication    & \textcolor{black} {-0.038}  &                      n/a   \\ \hline 
Google        & Dilation          &                     0.854   &                     0.824  \\ \hline
Google        & Ridge Regression  & \textcolor{blue}   {0.922}  & \textcolor{blue}   {0.829} \\ \hline
Google        & MLP               & \textcolor{magenta}{0.865}  & \textcolor{magenta}{0.827} \\ \hline
\hline
Mistral       & Simple Addition   &                     0.854   & \textcolor{magenta}{0.829} \\ \hline
Mistral       & Weighted Addition & \textcolor{magenta}{0.856}  & \textcolor{magenta}{0.829} \\ \hline
Mistral       & Multiplication    & \textcolor{black}  {-0.042}  &                      n/a   \\ \hline 
Mistral       & Dilation          & \textcolor{magenta}{0.856}  & \textcolor{magenta}{0.829} \\ \hline
Mistral       & Ridge Regression  & \textcolor{blue}   {0.914}  & \textcolor{blue}   {0.831} \\ \hline
Mistral       & MLP               &                     0.854   & \textcolor{magenta}{0.829} \\ \hline
\hline
OpenAI  & Simple Addition   &                     0.609   &                     0.819  \\ \hline
OpenAI  & Weighted Addition &                     0.609   &                     0.819  \\ \hline
OpenAI  & Multiplication    & \textcolor{black} { 0.001}  &                      n/a   \\ \hline 
OpenAI  & Dilation          &                     0.607   &                     0.818  \\ \hline
OpenAI  & Ridge Regression  & \textcolor{blue}   {0.767}  & \textcolor{blue}   {0.830} \\ \hline
OpenAI  & MLP               & \textcolor{magenta}{0.673}  & \textcolor{magenta}{0.825} \\ \hline
\hline
BERT          & Simple Addition   &                     0.891   &                     0.289  \\ \hline
BERT          & Weighted Addition &                     0.891   &                     0.294  \\ \hline
BERT          & Multiplication    & \textcolor{black} {-0.610}  &                      n/a   \\ \hline 
BERT          & Dilation          &                     0.891   &                     0.293  \\ \hline
BERT          & Ridge Regression  & \textcolor{blue}   {0.932}  & \textcolor{blue}   {0.501} \\ \hline
BERT          & MLP               & \textcolor{blue}   {0.932}  & \textcolor{magenta}{0.496} \\ \hline

\end{tabular}
\caption{Comparison of compositionality models for each type of embedding model, for the LADEC dataset. Numbers in \textcolor{blue}{blue} and \textcolor{magenta}{magenta} are the highest and second-highest for each embedding model.}
\label{table:main_results_table}
\end{table}

\begin{table}[htbp]
\centering
\begin{tabular}{|l|l|r|r|}
\hline
\textbf{Comp.} & \textbf{Dataset} & \textbf{Cosine} & \textbf{JS}   \\
\textbf{Model}            &                  & \textbf{Sim.}   & \textbf{Div.} \\ \hline
\hline
Simple Add.   & LADEC     &  0.854  & 0.829 \\ \hline
Simple Add.   & LADEC-NC  &  0.862  & 0.831 \\ \hline
\hline
Weighted Add. & LADEC     &  0.856  & 0.829 \\ \hline
Weighted Add. & LADEC-NC  &  0.863  & 0.832 \\ \hline
\hline
Multiplication    & LADEC     & -0.042  & n/a \\ \hline 
Multiplication    & LADEC-NC  & -0.037  & n/a \\ \hline 
\hline
Dilation          & LADEC     &  0.856  & 0.829 \\ \hline
Dilation          & LADEC-NC  &  0.863  & 0.832 \\ \hline
\hline
Ridge Reg.  & LADEC     &  0.914  & 0.831 \\ \hline
Ridge Reg.  & LADEC-NC  &  0.925  & 0.832 \\ \hline
\hline
MLP               & LADEC     &  0.854  & 0.828 \\ \hline
MLP               & LADEC-NC  &  0.827  & 0.831 \\ \hline

\end{tabular}
\caption{Comparing LADEC vs. LADEC-NC, existing and novel compounds show very similar compositionality metrics. Mistral embeddings are used here, but other embeddings give very similar results.}
\label{table:metrics_NC}
\end{table}


\section{Generalization to Systematic Compositions}

The LADEC dataset contains compound words like `bookstore', `dumbbell' and `waterfall', which are all single-word concepts composed from two constituent words. These compounds have varying levels of semantic transparency, and have arbitrary semantic relationships between the constituents. For example, `bookstore' is a ``store for books'', but `dumbbell' is not a ``bell for dumbs''. `Dumbbell' originates from it looking like a church bell, but is dumb or silent (i.e. the noun is the second constituent), whereas `waterfall' relates to water that falls (i.e. noun is first constituent).

Do the results in the previous section \textbf{generalize to two-word compositions that are composed more systematically}? To investigate this, we turn to adjectives and nouns, because in adjective-noun compositions, there is a clear semantic relationship between the constituents.

\subsection{Simple Adjective-Noun Compositions (SANC) Dataset}

We design a dataset that comprises all pairwise combinations of 25 common adjectives with 25 common nouns. The 25 adjectives are from 3 common categories (12 colors, 7 materials and 6 size/shape attributes). The full set of adjectives and nouns can be seen in Figure~\ref{fig:sem-dominance}.

Inspired by~\citet{baroni-zamparelli-2010-nouns} and~\citet{Gagne2019-qv}, we constrain our dataset to simple adjective-noun compositions, because common nouns and adjectives are exhaustively composable, meaning that any adjective can be combined with any noun, regardless of commonness or plausibility (e.g. \textit{metal papaya} or \textit{magenta watermelon}). Furthermore, the adjective-noun compositions are semantically transparent, i.e. \textit{red ball} is clearly both \textit{red} and a \textit{ball} (unlike \textit{flea market}, for instance).

\subsection{Methods and Experimental Setup}

For this section on generalization, we use only the classic \textbf{Simple Addition} compositionality model, due to its simplicity (no learning or parameters), and the fact that its performance is close to the best model (Ridge Regression).

\subsubsection{Semantically-Similar Phrases}

We want to gauge how the Simple Addition model fares in relation to semantically-similar phrases. For example, where $S(x, y)$ is the cosine similarity between $x$ and $y$, and \textless$p$\textgreater~is the embedding vector of phrase $p$, how does Simple Addition compare to semantically similar variations?

The comparisons include several formulations of $S$: (\textless red ball\textgreater, \textless red\textgreater~$+$~\textless ball\textgreater), (\textless red ball\textgreater, \textless red-colored ball\textgreater), (\textless red ball\textgreater, \textless ball is red-colored\textgreater), and (\textless red ball\textgreater, \textless color of ball is red\textgreater). Additionally, we evaluate against baselines such as (\textless red ball\textgreater, \textless red\textgreater), (\textless red ball\textgreater, \textless ball\textgreater), and (\textless red ball\textgreater, \textless color of door is green\textgreater).
From the SANC dataset, we cross the 12 color adjectives with all 25 nouns, then plot the KDEs (using $12*25=300$ compositions) separately for each variation or baseline (see Figure~\ref{fig:phrase_input}). BERT was not used, due to poor compositionality. We use only colors for this experiment, in order to automatically generate semantically-similar phrases (e.g. `red-colored ball').

\subsection{Results}

\subsubsection{Generalization to Adjective-Noun Compositions}
\label{sec:adj-noun-results}

\begin{figure}[htbp]
\centering
\includegraphics[width=\columnwidth]{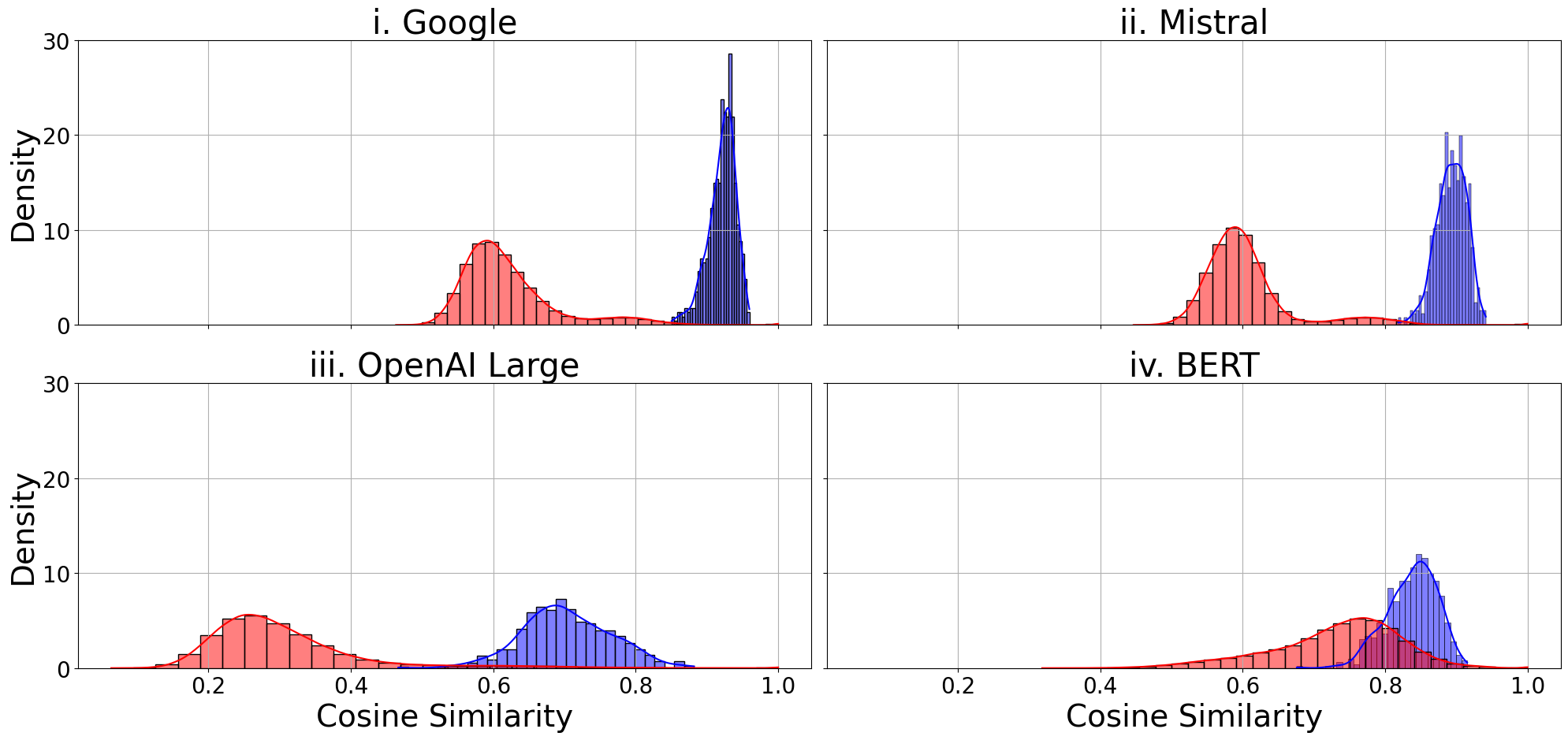}
\caption{KDEs across models for the SANC dataset. \textcolor{blue}{Blue plots} are the cosine similarity distributions using Simple Addition. \textcolor{red}{Red plots} are baseline distributions.}
\label{fig:simple_addition_SANC}
\end{figure}

From Figure~\ref{fig:simple_addition_SANC}, we see that the Simple Addition model also works well on the SANC dataset for Google, Mistral and OpenAI Large, with fairly high JS Divergence values and cosine similarities (details in Supplementary Materials). The good performances on LADEC, LADEC-NC and SANC point to the generality of the Simple Addition model across different types of compositions.

\subsubsection{Comparison to Semantically-Similar Phrases}

\begin{figure}[htbp]
\centering
\includegraphics[width=\columnwidth]{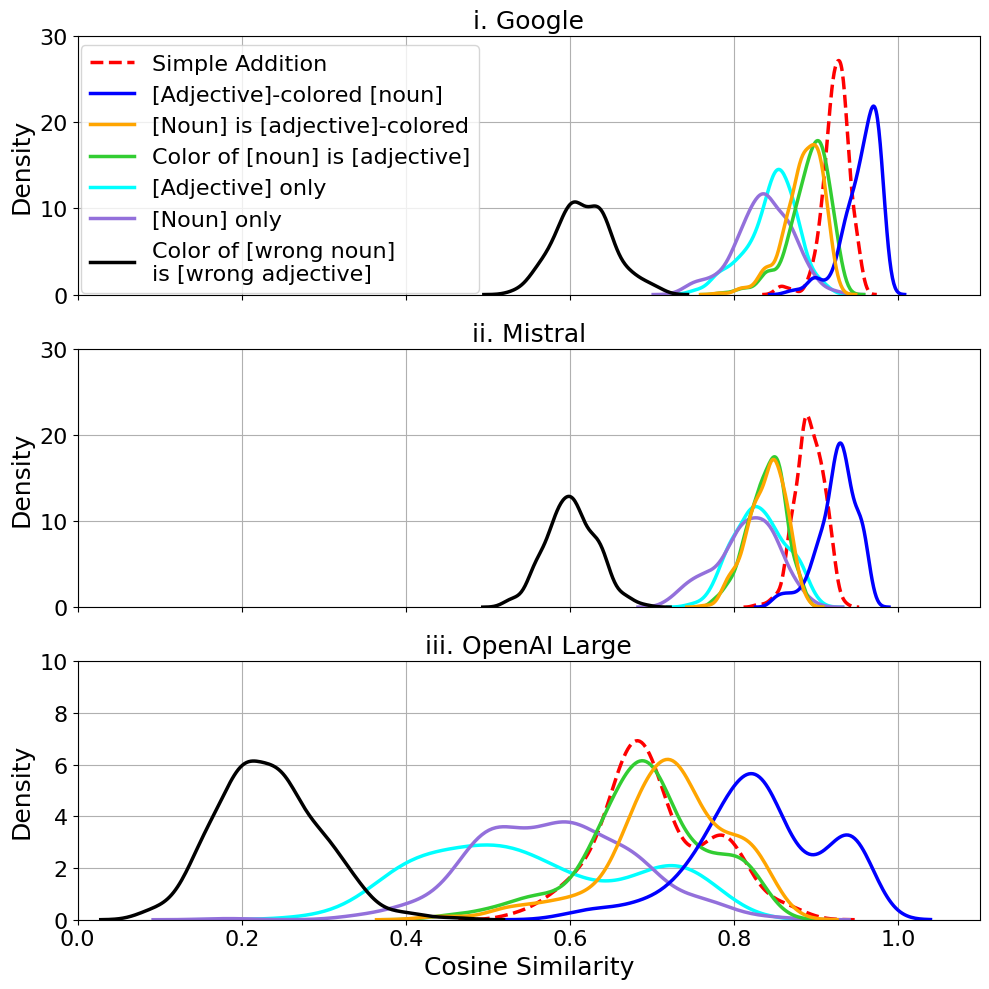}
\caption{Compositionality of Simple Addition in relation to 3 semantically-similar phrases and 3 baselines.}
\label{fig:phrase_input}
\end{figure}

From Figure~\ref{fig:phrase_input}, for both Google and Mistral, the order of the distributions (from highest to lowest) is roughly: blue, red, orange/green, cyan/purple, black. Using `red ball' to illustrate, this means that the similarity of the various embeddings to \textless red ball\textgreater, from most to least similar, are:
\vspace{-0.2cm}
\begin{itemize}
\item \textless red-colored ball\textgreater
\vspace{-0.35cm}
\item \textless red\textgreater~$+$~\textless ball\textgreater
\vspace{-0.35cm}
\item \textless ball is red-colored\textgreater~or~\textless color of ball is red\textgreater
\vspace{-0.35cm}
\item \textless red\textgreater~~or~~\textless ball\textgreater
\vspace{-0.35cm}
\item \textless color of door is green\textgreater
\end{itemize}
\vspace{-0.2cm}
These results not only match the rough intuition of how semantically similar the phrases are to `red ball', but show that the Simple Addition model \textless red\textgreater$+$\textless ball\textgreater~represents `red ball' almost as well as `red-colored ball' (which is semantically almost identical to `red ball').

Results for OpenAI Large are qualitatively similar, except that the red distribution is no longer distinctly higher than the orange and green ones.

\section{Deeper Performance Analysis}
\label{sec:performance-analysis}

Our findings suggest some insights for new models:

\begin{itemize}
    \item The strong performance of simple linear models suggests that transformer embeddings support linear additive compositionality. This aligns with previous findings and indicates that \textbf{embedding spaces could be optimized for linear operations}.
    \item BERT's poorer performance compared to models like Mistral and OpenAI suggests that \textbf{architectural choices} (e.g., bidirectional encoder vs. autoregressive models) and training objectives impact compositionality.
    \item Models can be trained with \textbf{optimization objectives} that promote linear separability and \textbf{semantic transparency}, potentially improving performance on downstream tasks requiring compositional understanding.
\end{itemize}

A complete discussion on this can be found in Appendix \ref{sec:deeper-analysis}.

\section{Qualitative Analyses}

\subsection{Methods}

\subsubsection{Constituent dominance} For each adjective-noun composition in SANC, we analyzed whether it is more dominated by (similar to) the adjective or the noun, i.e. we compare $S$ (\textless red ball\textgreater, \textless red\textgreater) to $S$ (\textless red ball\textgreater, \textless ball\textgreater). 

\subsubsection{Visualization using UMAP} We used UMAP -- Uniform Manifold Approximation and Projection~\cite{mcinnes2018umap} to visualize the embedding space of adjective-noun compositions in SANC.

\subsection{Results}
\label{sec:results}

\subsubsection{Constituent dominance} 
\begin{figure}[htbp]
\centering
\includegraphics[width=\columnwidth]{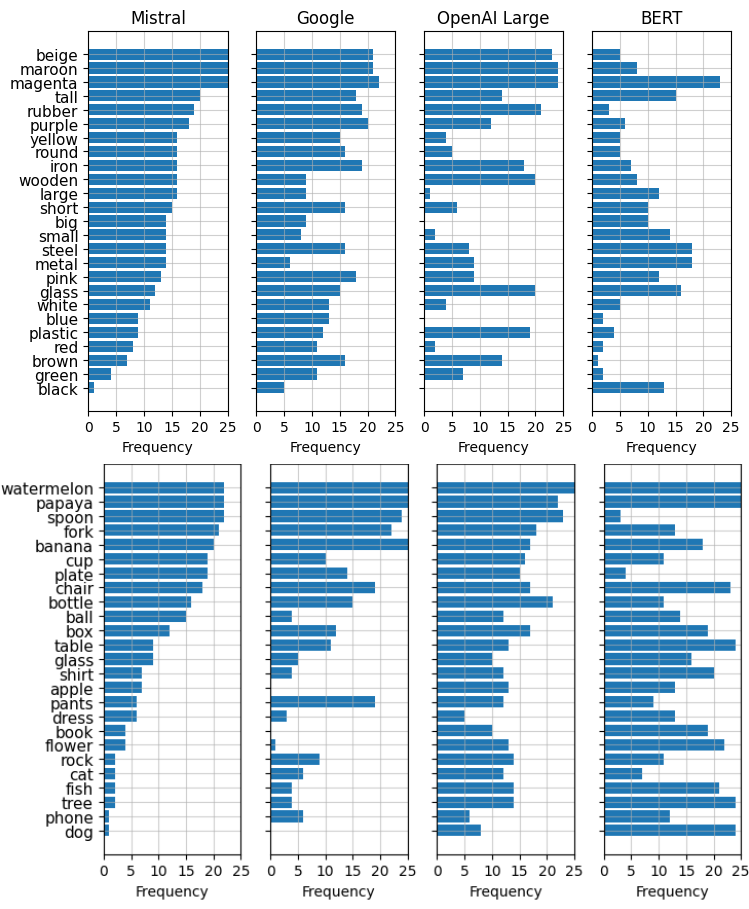}
\caption{Constituent dominance of \textbf{adjectives} and \textbf{nouns}, ordered by Mistral dominance rank.}
\label{fig:sem-dominance}
\end{figure}

From Figure~\ref{fig:sem-dominance}, we see that there is a diversity among adjective-dominance and noun-dominance; it is not the case the adjectives always dominate, or nouns always dominate. More importantly, we see that the embeddings are quite different, with Google, OpenAI Large and BERT increasingly different from Mistral. Despite these different representations, it is interesting that Google, Mistral and OpenAI Large give fairly similar results in Figure~\ref{fig:main_results_plots} across compositionality models.

\subsubsection{UMAP visualizations}

\begin{figure}[htbp]
\centering
\includegraphics[width=\columnwidth]{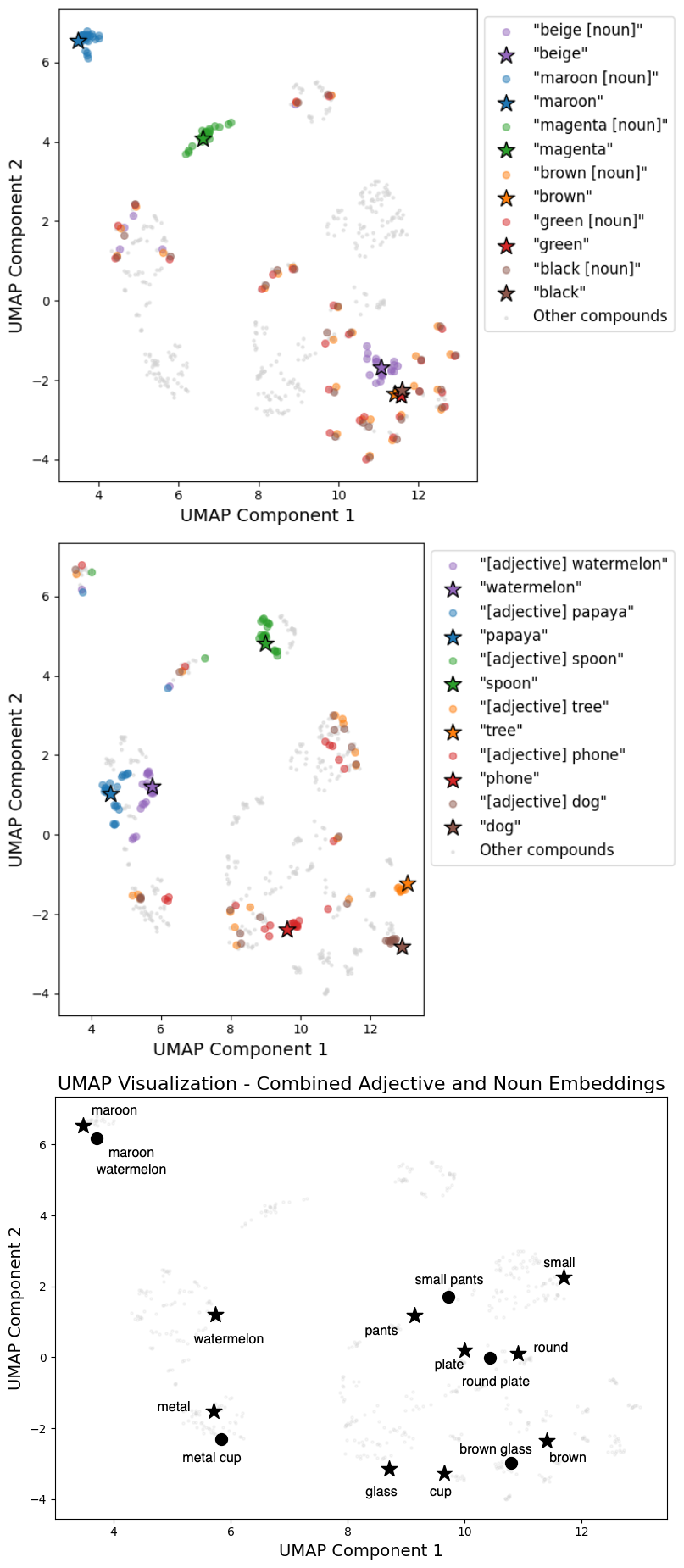}
\caption{UMAP plot of Mistral embeddings using Ridge Regression. The top figure shows the UMAP color-coded by \textbf{adjective}. The most dominant adjectives are `beige', `maroon' and `magenta'; least are `brown', `green' and `black'. The middle figures shows the UMAP colored by \textbf{noun}. The most dominant nouns are `watermelon', `papaya' and `spoon'; the least are `tree', `phone' and `dog'. The bottom-most UMAP plots the nouns, adjectives and their corresponding compositions on the same plot.
}
\label{fig:umap-noun}
\end{figure}

Figure~\ref{fig:umap-noun} (top and middle) shows the UMAP representations of Mistral embeddings predicted by the Ridge Regression model, labelled either adjective and noun constituent. We can see that when a constituent dominates the representation (e.g. `beige', `watermelon'), its compounds are well-clustered near it. However, for constituents that are not dominant (e.g. `brown', `tree'), its compounds are dispersed. It is important to note that that dominance is defined based on cosine similarity on the ground-truth embedding, while the visualization is based on UMAP (using euclidean distance) on predictions of Ridge Regression. These independent analyses suggest that Ridge Regression indeed performs semantic compositionality.

Figure~\ref{fig:umap-noun} (bottom) depicts a few example compositions (circles) and their constituents (stars). Qualitatively, it is interesting to note that several -- though not all -- compositions (e.g. `maroon watermelon', `metal cup', etc) are located between their constituents, consistent with linear models (such as Ridge Regression and Weighted Addition) being able to account for compositionality well.

These visualizations demonstrate that compound words tend to cluster around one of their constituent parts, confirming that semantic contributions within compounds are influenced by factors like semantic richness and contextual salience, rather than fixed syntactic rules \cite{libben1998semantic}. This finding aligns with the \textit{dual coding theory}, which highlights the deeper semantic processing of concrete nouns due to their imageability, leading to stronger clustering effects \cite{paivio1990mental}. Additionally, dominant adjectives in UMAP visualizations support the \textit{qualia structure} in generative lexicon theory, where adjectives play a crucial role in shaping the semantic interpretation of nouns \cite{pustejovsky1998generative}. The observed clustering behavior underscores the dynamic construction of compound meanings based on the semantic properties of their parts, resonating with \textit{contextualism} in semantic theory and \textit{semantic frame theory} \cite{recanati2004literal, fillmore2006frame}. 

\section{Conclusion}

In this study, we explored the compositionality of various transformer-based embedding models. For the first time, we comprehensively analyse modern embedding models and evaluate their compositionality over several metrics and factors. Our results showed that modern transformer models exhibit strong compositionality compared to BERT. Functional analysis revealed that ridge regression, a linear model, best explained compositionality in these embeddings. We also found that these embeddings are highly compositional regardless of whether the compound words were seen during training, suggesting that compositionality arises from the embedding models' architectures, rather than training process or data. 
Our work significantly advances our understanding of compositionality in modern language models.

\clearpage

\section{Limitations}
While our study provides valuable insights into the compositionality of transformer-based embeddings, several limitations must be acknowledged:

\begin{itemize}
    \item \textbf{Scope of Models:} We focused on a limited set of embedding models, specifically Google, Mistral, OpenAI Large, and BERT. Other prominent embeddings (e.g., GPT-4 embeddings or smaller, domain-specific models) were not included and may yield different results.

    \item \textbf{Simple Adjective-Noun Compositions:} Our experiments are constrained to two-word compositions, with an additional focus on adjective-noun pairs. This was done to control for syntactic complexity and precisely measure compositional properties. While this setup provides a controlled evaluation, it does not account for more complex compositional structures involving multiple modifiers or sentence-level contexts.

    \item \textbf{Dataset Bias:} The LADEC and SANC datasets, though effective for testing compositionality, may not fully capture real-world variability in compound words and phrases. Additionally, LADEC-NC consists of synthetically generated compounds, which may lack linguistic or cultural plausibility.

    \item \textbf{BERT-Specific Challenges:} BERT's poor performance, attributed to subword tokenization and bidirectionality, highlights specific architectural limitations. However, our analysis does not explore potential improvements for BERT (e.g., fine-tuning on compositional tasks) or the effects of alternative training strategies.

    \item \textbf{Linear Assumptions:} Our reliance on linear models (e.g., Ridge Regression and Simple Addition) assumes that compositionality can be approximated linearly. While this works well empirically, it may oversimplify the nuanced, non-linear interactions present in transformer embeddings.

    \item \textbf{Evaluation Metrics:} Cosine similarity and JS divergence were used as primary metrics to assess compositionality. While these metrics are widely used, they may not fully capture the semantic nuances or compositional relationships in all contexts.

    \item \textbf{Model Training Data:} We do not analyze the role of specific training data in enabling compositionality. The presence or absence of certain compounds during pretraining could influence the models' ability to generalize to novel compositions.

    \item \textbf{More models and architectures:} While we provide additional models in section \ref{sec: additional_models}, including more models and newer architectures can further improve the comprehensiveness of the findings.

    \item \textbf{Evaluation metrics:} additional semantic evaluation metrics beyond cosine similarity and JS divergence could provide richer insights into the compositionality nuances. We chose these metrics due to their interpretability and widespread acceptance in previous literature but exploring more precise and specific evaluation metrics could lead to richer understanding of the embedding representations.
\end{itemize}

Future work can address these limitations by including a broader range of embedding models, exploring complex multi-word compositions, and investigating the impact of alternative training objectives and evaluation frameworks.

\bibliography{acl_latex}

\clearpage

\appendix

\section{Appendix}

\subsection{Further Related works}
\label{sec:related-work-continued}
\textbf{Distributional Semantic Models (DSMs).} \citet{turney2010frequency} introduced DSMs, which derive vector-based word representations from patterns of word usage in large corpora. These models have proven effective for numerous applications, such as word similarity and analogy tasks. The seminal work by \citet{bullinaria2012extracting} highlighted the importance of pointwise mutual information (PMI) scores in constructing effective DSMs, laying the groundwork for subsequent models.

\subsection{Compute Details}
\begin{enumerate}
    \item The experiments were carried out on a compute server using an NVIDIA RTX 3060 with 12GB of VRAM, and an AMD Ryzen 5500 CPU with 16GB of RAM. For proprietary models, a standard API interface was used which has been provided in the code submisison. 
    \item \textbf{Software and libraries used.} Tensorflow \cite{tensorflow2015-whitepaper} was used for model loading and management. The complete set of modules, packages and software used is described in the code submission. 
    \item A random seed of 42 was used for all sampling purposes.
\end{enumerate}

\subsection{Adjective-Noun Compounds Dataset}
\label{adj-noun-dataset-details}

\begin{table}[ht!]
\caption{List of 25 adjectives and 25 nouns used for curating  adjective-noun compounds.}
\label{dataset}
\vskip 0.15in
\begin{center}
\begin{small}
\begin{sc}
\begin{tabular}{cc|cc}
\toprule
  \multicolumn{2}{c|}{\textbf{Adjectives}} & \multicolumn{2}{c}{\textbf{Nouns}}   \\
\midrule
beige & plastic & apple & fork  \\
big & purple & ball & glass  \\
black & red & banana & pants   \\
blue & round & book & papaya   \\
brown & rubber & bottle & phone   \\
glass & short & box & plate   \\
green & small & cat & rock   \\
iron & steel & chair & shirt   \\
large & tall & cup & spoon   \\
magenta & white & dog & table  \\
maroon & wooden & dress & tree  \\
metal & yellow & fish & watermelon  \\
pink &  & flower &  \\
\bottomrule
\end{tabular}
\end{sc}
\end{small}
\end{center}
\vskip -0.1in
\end{table}

\subsection{Ladec-NC Dataset}

The Ladec-NC dataset as described in the main paper is included in the supplemental in the file ladec\_nc.jsonl

\subsection{}{Mathematical Formulations for Techniques}
\label{composition_formulae}

\subsubsection{Simple Addition}
Simple Addition is a simple un-weighted addition of the constituent vector embeddings.
\[ p = \mathbf{u} + \mathbf{v} \]

\subsubsection{Additive Model}
The Additive Model sums the embeddings of the constituent words. It is expressed as:
\[ p = \alpha \mathbf{u} + \beta \mathbf{v} \]
where \( \alpha \) and \( \beta \) are scalar weights of the vectors \( \mathbf{u} \) and \( \mathbf{v} \).

\subsubsection{Multiplicative Model}
This model involves an element-wise multiplication of the embeddings:
\[ p_i = u_i v_i \]
for each component \( i \) of vectors \( \mathbf{u} \) and \( \mathbf{v} \).

\subsubsection{Dilation}
Dilation modifies the interaction between two vectors by stretching one towards another.
\[ p = (\mathbf{u} \cdot \mathbf{u})\mathbf{v} + (\lambda - 1)(\mathbf{u} \cdot \mathbf{v})\mathbf{u} \]
This model accounts for both direct and scaled vector interactions, where \( \lambda \) is a dilation parameter.

\subsection{Final Learned Parameters for Gradient Descent Based Parameter Learning in Additive and Dilation Methods}
\label{learned-params}
This appendix summarizes the final learned parameters for each model after training via gradient descent over 100 epochs. Each model employed either the additive or dilation composition method, as indicated.

\subsubsection*{Additive Gradient Descent Method}
\begin{itemize}
    \item \textbf{Google}: Weights = \([4.0680, 2.9924]\)
    \item \textbf{OpenAI Large}: Weights = \([4.3841, 3.9983]\)
    \item \textbf{Mistral}: Weights = \([5.3313, 3.9911]\)
    \item \textbf{BERT}: Weights = \([5.0875, 5.5244]\)
    \item \textbf{Mistral Novel Compound Words}: Weights = \([4.3603, 3.7515]\)
    \item \textbf{BERT Novel Compound Words}: Weights = \([5.4968, 4.7989]\)
\end{itemize}

\subsubsection*{Dilation Gradient Descent Method}
\begin{itemize}
    \item \textbf{Google}: \( \lambda \) = \( 3.1064 \)
    \item \textbf{OpenAI Large}: \( \lambda \) = \( 4.5377 \)
    \item \textbf{Mistral}: \( \lambda \) = \( 2.9842 \)
    \item \textbf{BERT}: \( \lambda \) = \( 2.0609 \)
    \item \textbf{Mistral Novel Compound Words}: \( \lambda \) = \( 2.8168 \)
    \item \textbf{BERT Novel Compound Words}: \( \lambda \) = \( 2.3679 \)
\end{itemize}

Each parameter vector or scalar \( \lambda \) was optimized to minimize the loss function specific to each model, reflecting how well the gradient descent approach could approximate the theoretical composition method.

\subsection{Grid Search Based Results for Additive and Dilation Models}
\label{grid_search_results}
\begin{figure*}
\centering
\begin{tabular}{cc}
  \includegraphics[width=0.4\linewidth]{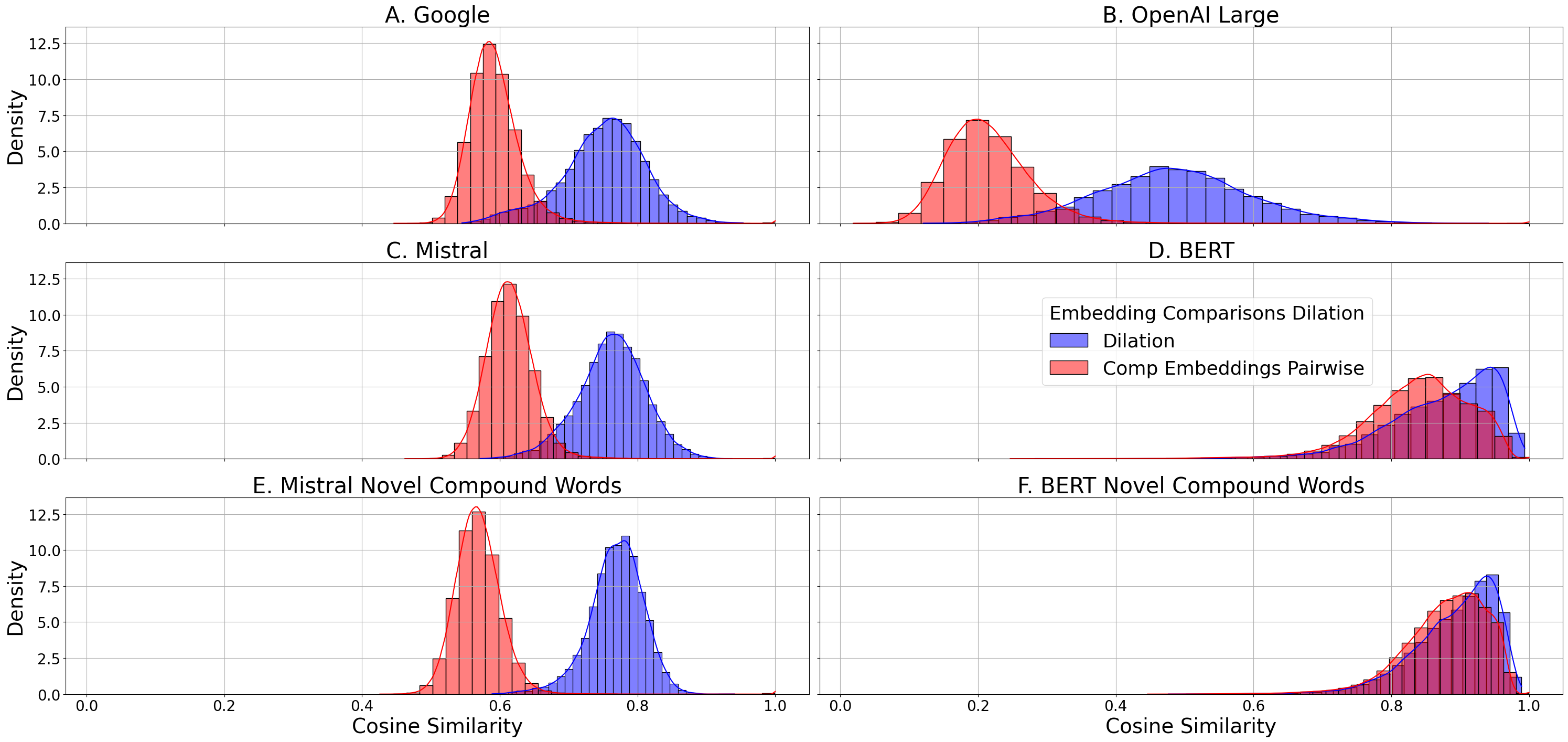} & 
  \includegraphics[width=0.4\linewidth]{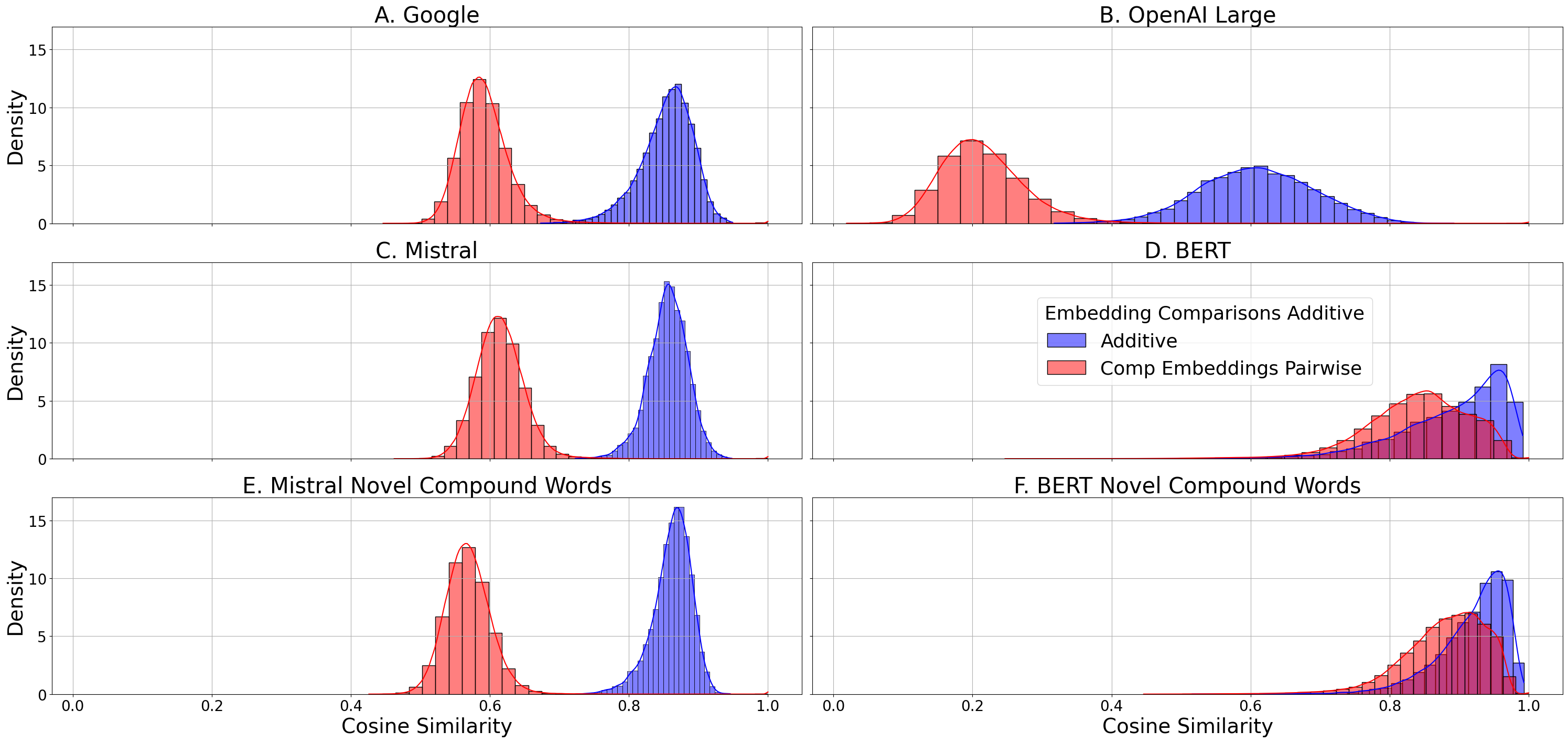} \\
  (a) Dilation - Grid Search & (b) Additive - Grid Search \\
\end{tabular}
\caption{Cosine similarity distribution densities across different composition models. Each subfigure shows the result from a different method or optimization approach, providing a comprehensive overview of the embedding model performances.}
\label{fig:compact_density_plots_additive_dilation_grid}
\end{figure*}

Table \ref{table:additive_dilation_metrics} provides grid search based results for Additive and Dilation Methods, and Figure \ref{fig:compact_density_plots_additive_dilation_grid} provides the corresponding Density plots. 

\begin{table}[htbp]
\centering
\caption{Metrics for Additive and Dilation Methods Across Different Models}
\label{table:additive_dilation_metrics}
\begin{tabular}{|p{2cm}|p{1.5cm}|p{1cm}|p{1cm}|}
\hline
\textbf{Model} & \textbf{Method} & \textbf{Cos. Dist.} & \textbf{JS Div.} \\ \hline
Google & Additive & 0.1460 & 0.8242 \\ \hline
OpenAI Large & Additive & 0.3910 & 0.8190 \\ \hline
Mistral & Additive & 0.1442 & 0.8289 \\ \hline
BERT & Additive & 0.1087 & 0.2893 \\ \hline
Mistral Novel Compound Words & Additive & 0.1381 & 0.8314 \\ \hline
BERT Novel Compound Words & Additive & 0.08055 & 0.2744 \\ \hline
Google & Dilation & 0.2493 & 0.7225 \\ \hline
OpenAI Large & Dilation & 0.5148 & 0.7287 \\ \hline
Mistral & Dilation & 0.2381 & 0.7524 \\ \hline
BERT & Dilation & 0.1252 & 0.2088 \\ \hline
Mistral Novel Compound Words & Dilation & 0.2309 & 0.8134 \\ \hline
BERT Novel Compound Words & Dilation & 0.1073 & 0.1036 \\ \hline
\end{tabular}
\end{table}

\subsubsection{Optimal Parameters from Grid Search}

In our grid search for tuning the model parameters, we explored a range of weight combinations for the Additive method and a set of lambda values for the Dilation method. The weight combinations tested included values such as (0.5, 0.5), (0.6, 0.4), (0.2,0.8), (0.1,0.9) and others, while the lambda values ranged from -1 to 1. 

The following table summarizes the final parameters selected for each model after the completion of the grid search process:

\begin{table}[htbp]
\centering
\caption{Final Parameters Selected from Grid Search}
\label{table:final_grid_search_params}
\begin{tabular}{|l|p{1.2cm}|p{1.2cm}|}
\hline
\textbf{Model} & \textbf{Additive Weights} & \textbf{Dilation Lambda} \\ \hline
Google & (0.6, 0.4) & 1 \\ \hline
OpenAI Large & (0.5, 0.5) & 1 \\ \hline
Mistral & (0.6, 0.4) & 1 \\ \hline
BERT & (0.5, 0.5) & 1 \\ \hline
Mistral Wrong Phrases & (0.5, 0.5) & 1 \\ \hline
BERT Wrong Phrases & (0.5, 0.5) & 1 \\ \hline
\end{tabular}
\end{table}

These parameters were selected based on their performance in optimizing the respective loss functions used in each model's evaluation, aiming to achieve the best results for the compositionality of embeddings in both the Additive and Dilation composition methods. An alternate Grid search was also performed for dilation where the lambda could take up to a max value of 40, at intervals of 1. The grid search suggested that lambda always optimized to the maximum possible value of available grid parameters (ie, 20 if the max value was 20 and 40 if the grid space ranged till 40 and so on). The density plots and metrics in all cases was similar to the provided results.

\subsection{Correlation of compound embedding cosine similarities with Semantic Metrics}
Tables \ref{tab:google_corr} and \ref{tab:openai_large_corr} provide the complete correlation tables for the compound embeddings created by Google Embeddings and OpenAI Large Embeddings and their correlation with semantic metrics provided in the LADEC Dataset.

Compositionality is the dataset’s meaning-predictability score (\texttt{ratingcmp}); it shows how easily a reader can guess a compound’s sense from its two parts.  
\texttt{Juhasz\_tran} repeats semantic-transparency ratings collected by Juhasz, Lai and Woodcock for 629 English compounds \cite{juhasz2015database, Juhasz2015}.  
\texttt{ratingC1} and \texttt{ratingC2} give meaning-retention judgments for the first and second constituents; \texttt{st\_c1\_mean} and \texttt{st\_c2\_mean} come from an update by \citet{Kim2019}.  
The three LSA values—\texttt{LSAc1stim}, \texttt{LSAc2stim}, \texttt{LSAc1c2}—are cosine similarities from Latent Semantic Analysis: first part versus whole compound, second part versus whole compound, and the two parts together \cite{Landauer1997}.  
Concreteness scores (\texttt{concreteness\_stim}, \texttt{concreteness\_c1}, \texttt{concreteness\_c2}) originate in the Brysbaert et al.\ 40 k norms; higher numbers indicate richer perceptual content \cite{Brysbaert2009, Brysbaert2014}. 
Valence scores (\texttt{valence\_stim}, \texttt{valence\_c1}, \texttt{valence\_c2}) come from Warriner et al.\ affective ratings; larger values point to a more pleasant tone \cite{Warriner2013}.  
Sentiment variables (\texttt{sentimentprobpos\_}, \texttt{sentimentprobneg\_}, \texttt{sentimentratioposneg\_}) are outputs of a Mathematica sentiment classifier for the compound (\texttt{\_stim}) and for each part (\texttt{\_c1}, \texttt{\_c2}).  
\texttt{Zipfvalue} presents frequency on the 1–7 Zipf scale (log\textsubscript{10} occurrences per billion words + 3) \cite{vanHeuven2014}.  
\texttt{stim\_SLlg10wf}, \texttt{c1\_SLlg10wf} and \texttt{c2\_SLlg10wf} store SUBTLEX-US log\textsubscript{10} word-frequency values for the compound and its two constituents \cite{Brysbaert2009}.

While the correlation values (Pearson) reported in Tables 6 and 7 are moderate, they are positive for well-performing models like Google and Mistral across key semantic factors—such as the predictability of the compound from its constituents (semantic transparency), meaning retentions of constituent 1 and constituent 2, and relatedness of the compound to each constituent. In contrast, BERT shows correlations that are close to zero or even negative for some metrics.

Note that the correlation values are always higher for the head (constituent 2) than the modifier (constituent 1). This observation aligns with human ratings. In LADEC, the mean normalized meaning retention based on human scores for constituent 2 is 0.693, while for constituent 1 it is 0.644. This appears to reflect the natural behavior of English compounds, where the head typically carries more semantic weight, and the models appear to capture this tendency. Our deeper analysis through the UMAP projections in section \ref{sec:results} dives deeper into this phenomenon through examination of the existing linguistic theories around this subject.

Tables \ref{tab:google_corr} and \ref{tab:openai_large_corr} show positive correlations between embedding-based compositionality scores and human-annotated semantic transparency ratings. These findings confirm our expectation: transparent compounds ("red ball") exhibit high similarity to the sum of their constituents’ embeddings ("red" + "ball"), whereas opaque compounds ("bluetooth") do not. Thus, our results demonstrate that the models behave consistently with human semantic judgments.


\begin{table}[ht!]
\centering
\scriptsize
\begin{tabular}{|l|r|}
\hline
\textbf{Metric} & \textbf{Correlation} \\ \hline
compositionality & 1.0000 \\ \hline
ratingC2 & 0.4352 \\ \hline
Juhasz\_tran & 0.3888 \\ \hline
st\_c2\_mean & 0.3218 \\ \hline
LSAc2stim & 0.2868 \\ \hline
ratingcmp & 0.2804 \\ \hline
st\_c1\_mean & 0.2398 \\ \hline
LSAc1stim & 0.2282 \\ \hline
concreteness\_c2 & 0.2194 \\ \hline
LSAc1c2 & 0.1676 \\ \hline
ratingC1 & 0.1616 \\ \hline
concreteness\_stim & 0.1360 \\ \hline
concreteness\_c1 & 0.08303 \\ \hline
c2\_SLlg10wf & 0.07585 \\ \hline
valence\_stim & 0.06658 \\ \hline
sentimentratioposneg\_c2 & 0.05392 \\ \hline
valence\_c2 & 0.04755 \\ \hline
sentimentprobneg\_stim & 0.04322 \\ \hline
sentimentprobpos\_c1 & 0.01612 \\ \hline
sentimentprobpos\_c2 & 0.01163 \\ \hline
sentimentratioposneg\_c1 & 0.007423 \\ \hline
sentimentprobneg\_c1 & -0.01365 \\ \hline
sentimentprobpos\_stim & -0.01573 \\ \hline
valence\_c1 & -0.03590 \\ \hline
sentimentprobneg\_c2 & -0.03883 \\ \hline
sentimentratioposneg\_stim & -0.04266 \\ \hline
c1\_SLlg10wf & -0.06150 \\ \hline
Zipfvalue & -0.07600 \\ \hline
stim\_SLlg10wf & -0.07600 \\ \hline
\end{tabular}
\caption{Correlation values for Google model.}
\label{tab:google_corr}
\end{table}

\begin{table}[ht!]
\centering
\scriptsize
\begin{tabular}{|l|r|}
\hline
\textbf{Metric} & \textbf{Correlation} \\ \hline
compositionality & 1.0000 \\ \hline
Juhasz\_tran & 0.3449 \\ \hline
ratingC2 & 0.2851 \\ \hline
st\_c2\_mean & 0.2823 \\ \hline
st\_c1\_mean & 0.2713 \\ \hline
LSAc1stim & 0.2284 \\ \hline
ratingcmp & 0.2279 \\ \hline
LSAc2stim & 0.2104 \\ \hline
LSAc1c2 & 0.1658 \\ \hline
concreteness\_stim & 0.1639 \\ \hline
ratingC1 & 0.1596 \\ \hline
valence\_stim & 0.1375 \\ \hline
concreteness\_c2 & 0.1099 \\ \hline
concreteness\_c1 & 0.08663 \\ \hline
sentimentratioposneg\_c2 & 0.03800 \\ \hline
sentimentratioposneg\_c1 & 0.02216 \\ \hline
sentimentprobpos\_c1 & 0.02207 \\ \hline
sentimentprobpos\_c2 & 0.01885 \\ \hline
sentimentratioposneg\_stim & 0.008305 \\ \hline
sentimentprobpos\_stim & 0.008204 \\ \hline
valence\_c2 & -0.002929 \\ \hline
sentimentprobneg\_stim & -0.005037 \\ \hline
sentimentprobneg\_c2 & -0.01864 \\ \hline
sentimentprobneg\_c1 & -0.02530 \\ \hline
valence\_c1 & -0.04324 \\ \hline
Zipfvalue & -0.05018 \\ \hline
stim\_SLlg10wf & -0.05018 \\ \hline
c2\_SLlg10wf & -0.1207 \\ \hline
c1\_SLlg10wf & -0.1856 \\ \hline
\end{tabular}
\caption{Correlation values for OpenAI Large model.}
\label{tab:openai_large_corr}
\end{table}

\begin{table}[ht!]
\centering
\scriptsize
\begin{tabular}{|l|r|}
\hline
\textbf{Metric} & \textbf{Correlation} \\ \hline
compositionality & 1.0000 \\ \hline
ratingC2 & 0.3627 \\ \hline
LSAc2stim & 0.3193 \\ \hline
st\_c2\_mean & 0.3095 \\ \hline
Juhasz\_tran & 0.2553 \\ \hline
ratingcmp & 0.2238 \\ \hline
concreteness\_stim & 0.2232 \\ \hline
LSAc1stim & 0.2120 \\ \hline
concreteness\_c2 & 0.1699 \\ \hline
st\_c1\_mean & 0.1599 \\ \hline
LSAc1c2 & 0.1503 \\ \hline
ratingC1 & 0.08918 \\ \hline
valence\_stim & 0.08348 \\ \hline
concreteness\_c1 & 0.05163 \\ \hline
stim\_SLlg10wf & 0.04702 \\ \hline
Zipfvalue & 0.04702 \\ \hline
sentimentprobneg\_c1 & 0.01716 \\ \hline
sentimentprobneg\_c2 & 0.01496 \\ \hline
sentimentratioposneg\_c2 & 0.01179 \\ \hline
sentimentprobpos\_c1 & 0.002194 \\ \hline
sentimentprobneg\_stim & -0.003203 \\ \hline
sentimentratioposneg\_stim & -0.003465 \\ \hline
sentimentprobpos\_stim & -0.009310 \\ \hline
valence\_c2 & -0.01035 \\ \hline
sentimentratioposneg\_c1 & -0.01967 \\ \hline
sentimentprobpos\_c2 & -0.02335 \\ \hline
c2\_SLlg10wf & -0.03055 \\ \hline
valence\_c1 & -0.07380 \\ \hline
c1\_SLlg10wf & -0.1286 \\ \hline
\end{tabular}
\caption{Correlation values for Mistral model.}
\label{tab:mistral_corr}
\end{table}

\begin{table}[ht!]
\centering
\scriptsize
\begin{tabular}{|l|r|}
\hline
\textbf{Metric} & \textbf{Correlation} \\ \hline
compositionality & 1.0000 \\ \hline
c2\_SLlg10wf & 0.1528 \\ \hline
Zipfvalue & 0.1195 \\ \hline
stim\_SLlg10wf & 0.1195 \\ \hline
valence\_stim & 0.07592 \\ \hline
valence\_c2 & 0.06451 \\ \hline
sentimentprobpos\_c1 & 0.04225 \\ \hline
sentimentratioposneg\_stim & 0.03315 \\ \hline
sentimentratioposneg\_c1 & 0.02905 \\ \hline
sentimentprobneg\_c2 & 0.01631 \\ \hline
st\_c1\_mean & 0.008733 \\ \hline
valence\_c1 & 0.007264 \\ \hline
sentimentprobpos\_stim & 0.006798 \\ \hline
sentimentratioposneg\_c2 & 0.003231 \\ \hline
sentimentprobpos\_c2 & 0.002738 \\ \hline
concreteness\_c2 & 0.002198 \\ \hline
c1\_SLlg10wf & -0.003704 \\ \hline
LSAc1c2 & -0.02588 \\ \hline
LSAc1stim & -0.02968 \\ \hline
sentimentprobneg\_c1 & -0.03122 \\ \hline
LSAc2stim & -0.03632 \\ \hline
sentimentprobneg\_stim & -0.04079 \\ \hline
concreteness\_c1 & -0.04906 \\ \hline
Juhasz\_tran & -0.05242 \\ \hline
ratingC2 & -0.05843 \\ \hline
st\_c2\_mean & -0.06931 \\ \hline
ratingC1 & -0.07454 \\ \hline
ratingcmp & -0.1124 \\ \hline
concreteness\_stim & -0.1471 \\ \hline
\end{tabular}
\caption{Correlation values for BERT model.}
\label{tab:bert_corr}
\end{table}

\subsection{SANC Simple Addition Numerical Results}
\begin{table}[H]
\centering
\begin{tabular}{|l|r|r|}
\hline
\textbf{Embedding Model} & \textbf{Cos. Sim.} & \textbf{JS Div.} \\ \hline
\hline
Google         &               0.921  & 0.824 \\ \hline
Mistral        &               0.893  & 0.822 \\ \hline
OpenAI Large   &               0.703  & 0.779 \\ \hline
BERT           &               0.838  & 0.542 \\ \hline
\end{tabular}
\caption{Metrics for Simple Addition on the SANC dataset.}
\label{table:simple_addition_SANC}
\end{table}

\subsection{Additional Models}
\label{sec: additional_models}

While our main analysis included four embedding models (Google, OpenAI, Mistral, and BERT), we also conducted preliminary evaluations on several additional smaller-scale models and other architectures. We provide JS divergence results for these models and architectures on the SANC dataset using the additive method in table \ref{table: additional_models}, which align with our previous findings.

\begin{table}[h]
    \centering
    \begin{tabular}{|l | c |}
        \hline
        Model & JS Divergence \\
        \hline
        OpenAI Small & 0.770269 \\
        CLIP         & 0.809656 \\
        FastText     & 0.555278 \\
        Comp\textendash CLIP & 0.808932 \\
        \hline
    \end{tabular}
    \caption{Performance of Additional models on SANC dataset and additive method}
    \label{table: additional_models}
\end{table}

\section{Result Analysis}
\label{sec:deeper-analysis}
Our findings suggest some insights for new models:

\begin{itemize}
    \item The strong performance of simple linear models suggests that transformer embeddings support linear additive compositionality. This aligns with previous findings and indicates that \textbf{embedding spaces could be optimized for linear operations}.
    \item BERT's poorer performance compared to models like Mistral and OpenAI suggests that \textbf{architectural choices} (e.g., bidirectional encoder vs. autoregressive models) and training objectives impact compositionality.
    \item Models can be trained with \textbf{optimization objectives} that promote linear separability and \textbf{semantic transparency}, potentially improving performance on downstream tasks requiring compositional understanding.
\end{itemize}

The poor performance of BERT can be explained by the following factors:

\subsection*{Bidirectionality in architecture:}
\begin{itemize}
    \item BERT's bidirectional nature means that embeddings are context-dependent and influenced by both left and right contexts.
    \item This can lead to embeddings that are less stable and less amenable to simple additive composition, as they may capture complex interactions that are not easily linearized.
\end{itemize}

\subsection*{Masked Language Modeling (MLM) as training objective:}
\begin{itemize}
    \item BERT is trained using MLM, where tokens are masked, and the model predicts them based on context.
    \item This training may not encourage the model to learn representations that facilitate linear compositionality, as the focus is on contextual prediction rather than encoding compositional semantics.
\end{itemize}

\subsection*{Subword Tokenization in BERT:}
\begin{itemize}
    \item BERT's WordPiece tokenizer consistently splits words into subword units, which can fragment compound words.
    \item This fragmentation may result in embeddings that don't capture the full semantic meaning of the compound word.
    \item For example, a novel compound like ``deepfake'' might be split into ``deep'' and ``\#\#fake,'' and the model might not have sufficient context to combine these subwords effectively.
    \item This can affect BERT's ability to generate meaningful embeddings for compound words, hindering compositionality.
    \item The other embedding models are also trained on a larger number of tokens which means a lot more compound words and complex phrases are natively present in their embedding representations.
\end{itemize}

\subsection*{Difference in Scale:}
\begin{itemize}
    \item The difference in the model size of BERT and the other embedding models could also be a potential factor in the relatively poor embedding quality for BERT.
\end{itemize}

In contrast, models like Mistral and OpenAI embeddings may be trained with objectives or architectures that better preserve compositional properties, or they might utilize tokenization methods that maintain whole-word representations more effectively.

\section*{Presence of Target Compounds in Model Vocabularies and Tokenization Effects}
An important point to note in this analysis is the impact of tokenization and presence or absence of target compounds from the model vocabulary for each of the models. While intuitively one might assume that compounds not in the model vocabulary might be split into subparts which could ultimately yield some sort of weighted average, in practice this is not necessarily true.

The highly non-linear and multi-layered nature of these models makes the representation and the impact of tokenization more ambiguous than trivial—A compound represented as a single token (e.g., “bluetooth”) may still be mapped to a compositional meaning internally, while two separate tokens (“blue” + “tooth”) may be jointly mapped to a distinct, possibly non-compositional, representation.
The crucial point to emphasize is that here is that even when a compound like “bluetooth” is tokenized into “blue” and “tooth,” transformer models do not necessarily represent the meaning of the compound as a weighted sum of its token embeddings as the implicit composition function.
These models may learn that specific token sequences correspond to non-compositional meanings during their training phase and encode this context in their representation phase. For instance, in pretraining data, "blue" and "tooth" might co-occur frequently in technological contexts, and the model may associate the entire phrase with a concept of technology rather than just with either word alone.

This variability makes it difficult to predict how meaning is constructed, and highlights why analyzing whether the relevance of composition functions, linguistic theories, and priors from VSM studies for transformer-based embeddings becomes even more important. This sort of emergent behavior has been documented in recent interpretability studies \cite{templeton2024scaling}. 
Such tokenization makes the composition task less trivial for the model, warranting an investigation into composition representation of such transformer models. This is especially true when we account for the fact that compound words may appear in multiple forms (e.g., “bluetooth” vs. “blue tooth”) in the training corpora, where their meanings shift across contexts and impact model representations \cite{haber-poesio-2024-polysemy, grindrod2024transformers}. 
The impact of tokenization on such polysemous constituents or compounds can introduce ambiguity in compositional interpretation, particularly when the model’s pretraining fails to capture dominant or contextually appropriate senses—an effect compounded by recent findings on superposition \cite{templeton2024scaling}, where overlapping semantic features are encoded in shared subspaces, making disambiguation more difficult.

\end{document}